\documentclass{article}

\usepackage{microtype}
\usepackage{graphicx}
\usepackage{subfigure}
\usepackage{booktabs} 
\usepackage{amsmath}
\usepackage{amsfonts}
\usepackage{physics}
\usepackage{bm}
\usepackage{cite}
\usepackage[round]{natbib} 

\usepackage{hyperref}

\newcommand{\lcd}{\mathcal{L}_{\text{CD}}}
\newcommand{\lkl}{\mathcal{L}_{\text{KL}}}
\newcommand{\lent}{\mathcal{L}_{\text{ent}}}
\newcommand{\lopt}{\mathcal{L}_{\text{opt}}}
\newcommand{\lcda}{\mathcal{L}_{\text{CD*}}}
\newcommand{\lnll}{\mathcal{L}_{\text{NLL}}}
\newcommand{\dkl}{D_{\text{KL}}}

\usepackage[accepted]{icml2018_ift6269}

\icmltitlerunning{Clarifying MCMC-based training of modern EBMs : Contrastive Divergence versus Maximum Likelihood}

\begin{document}

\twocolumn[
\icmltitle{Clarifying MCMC-based training of modern EBMs :\\ Contrastive Divergence versus Maximum Likelihood}



\icmlsetsymbol{equal}{*}

\begin{icmlauthorlist}
\icmlauthor{Léo Gagnon}{mila}
\icmlauthor{Guillaume Lajoie}{mila}
\end{icmlauthorlist}

\icmlcorrespondingauthor{Léo Gagnon}{leo.gagnon@mila.quebec}
\icmlaffiliation{mila}{Mila, Université de Montréal}


\vskip 0.3in
]
\printAffiliationsAndNotice{}




\begin{abstract}
The Energy-Based Model (EBM) framework is a very general approach to generative modeling that tries to learn and exploit probability distributions only defined though unnormalized scores. It has risen in popularity recently thanks to the impressive results obtained in image generation by parameterizing the distribution with Convolutional Neural Networks (CNN). However, the motivation and theoretical foundations behind modern EBMs are often absent from recent papers and this sometimes results in some confusion. In particular, the theoretical justifications behind the popular MCMC-based learning algorithm Contrastive Divergence (CD) are often glossed over and we find that this leads to theoretical errors in recent influential papers \cite{du2020,du2021}. After offering a first-principles introduction of MCMC-based training, we argue that the learning algorithm they use can in fact not be described as CD and reinterpret theirs methods in light of a new interpretation. Finally, we discuss the implications of our new interpretation and provide some illustrative experiments.
\end{abstract}
\vspace{-1cm}

\section{Introduction}
\label{intro}

Learning statistical models of observational data has always been one of the primary goals in machine learning and more generally in statistics. With such a model in hand, one can understand how the data is distributed and---of particular importance for this work---generate new samples as if they were coming from the data distribution.

To learn such a model, the first step is always to hypothesize a parametric family of distributions $p_\theta(x)$ which we think encompasses the real distribution---or a good approximation of it. Then we can use various statistical tools to infer which $\theta \in \Theta$ fits the data the best. 

Any choice of parametric family can be written as $$p_\theta(x)=\frac{1}{Z_\theta}\tilde{p}_\theta(x)$$ where $\tilde{p}_\theta\geq0$ is the \textit{unnormalized probability} and $Z_\theta=\int \tilde{p}_\theta(x) dx$ is the \textit{normalization constant}. Energy-based models (EBMs) consist in paremeterizing the family by defining the unnormalized probability in the following way
$$p_\theta(x)=\frac{1}{Z_\theta}e^{-E_\theta(x)}$$
where $E_\theta(x)\geq 0$ is called the \textit{energy function} and associates a scalar measure of \textit{incompatibility} to every point. The particular form of this distribution comes from statistical physics and is explained in section \ref{physics}. The main advantage of EBMs is to give the most flexibility possible to how we parameterize the parametric family : $E_\theta(x)$ can be any function and does not need to be normalized. In particular, recent deep learning work have exploited that to model image distribution by parametrizing $E_\theta(x)$ with a CNN : the ultimate goal being to generate new images coming from the learned distribution. 

However, we notice in these papers a growing confusion related to the theoretical underpinnings of the MCMC-based training procedure. In particular, we argue that two recent papers \cite{du2020,du2021} wrongly identify the learning algorithm they use as Contrastive Divergence, learning to confusion and unforeseen consequences. More precisely, in this paper :
\begin{itemize}
    \item We offer a much needed first-principles introduction to EBM sampling and training with MCMC-based objectives : Negative Log Likelihood (NLL) and Contrastive Divergence (CD). 
    \item We perform a detailed critique of two recent paper \cite{du2020,du2021} and argue that objective they are using is NLL while they are claiming it is CD. Moreover, we identify a mathematical error in the derivation of an auxiliary loss in \citep{du2021} and show how, with our new interpretation of their learning algorithm, it ends up being a good thing.
    \item We perform some simple experiments on MNIST to illustrate our main arguments.
\end{itemize}

\section{Physics of EBMs} 
\label{physics}
\subsection{The Boltzmann-Gibbs distribution}
\label{boltzmann}

Suppose that we have a system $S$ which is described by a state $x\in X$ and that this system is rapidly fluctuating due to random interactions. Next suppose that every state is associated with an energy function $E(x)$ and that we know that \textit{in expectation}, when we observe the system, it is going to have energy $\langle E \rangle<\infty$. In the language of physics, the second assumption is that the system is at \textit{thermal equilibrium} with a \textit{thermal bath at a fixed temperature} (i.e. $S$ interacts with another system that maintains it a fixed average energy). Then, the Boltzmann-Gibbs distribution gives us the probability that we observe the system $S$ in state $x$ :
$$p(x) = \frac{1}{Z}e^{-E(x)/T}$$
Where $T$ is the temperature and is related to $\langle E \rangle$. This tells us that the likelihood of a state decreases exponentially when its energy goes up : low energy states are preferred. It turns out that $p(x)$ can be derived as the MaxENT distribution with constraint $\mathbb{E}_{x\sim p}[E(x)]=\langle E\rangle$ \cite{jaynes57}. We can therefore understand $p(x)$ as the natural tendency of the system given that no states are a priori preferred (i.e. under the second law of thermodynamics). See \citealp{physicsofebms} for a thorough discussion of physics and EBMs.

Importantly, notice that what we have said here is equivalent to saying that the stationary distribution of states visited by a certain trajectory of the system (in contact with a thermal bath at a fixed temperature) is the Boltzmann-Gibbs distribution : if you start the system in state $x_0$ and let it run to infinity, it will visit visit each state $x$ in proportion to $p(x)$.

\subsection{EBMs as a physical simulation}
\label{simulation}

Following on this last point, consider the effect of doing MCMC sampling on a EBM with energy function $E_\theta(x)$. By definition, if the MCMC algorithm is valid, it will have as stationary distribution $p_\theta(x)=\frac{1}{Z_\theta}e^{-E_\theta(x)}$. But this is exactly the stationary distribution that a simulation an hypothetical system with energy $E_\theta(x)$ would have!

This means that we can think of running a MCMC chain on an EBM (i.e. sampling) as just simulating the hypothetical system defined by $E_\theta(x)$ in contact with a heat bath at a fixed temperature (in fact we suppose that $T=1$ for the rest of the paper).

Actually, the MCMC technique that we are going to use (section \ref{langevin}) is explicitly the simulation of a physical system. 

\subsection{Langevin sampling}
\label{langevin}

Because our ultimate objective is to parametrize the energy function with a neural network, we want a MCMC sampling algorithm that is general (unlike Gibbs sampling which relies on knowing conditional densities of input dimensions) and efficient (unlike random-walk Metropolis-Hastings which has very low acceptance probability in high dimension). The generally adopted strategy, \textit{Langevin sampling}, is the discretisation of a differential equation which describes the motion of a certain physical system. 

\subsubsection{Pollen in a glass of water}
\label{pollen}

Suppose you drop a grain of pollen in a glass of still water. First, because the particle wants to reduce its gravitational potential energy $U(x)$, it is going to move toward the bottom of the cup. However, because it is subjected to random molecular bombardment, a random component will be added to its trajectory. Because of the discussion in [section 2.1], it should make sense that the distribution of the particle's position follows the Boltzmann-Gibbs distribution $p(x)=\frac{1}{Z}e^{-U(x)}$. Equivalently, the stationary distribution of the differential equation describing the motion of the pollen grain (the Langevin equation) is the Boltzmann-Gibbs distribution.

Importantly, we can extrapolate this simple example to any potential we want : in our case $E_\theta(x)$.

\subsubsection{Discretization}
\label{discrete}

\begin{figure}[H]
\centering
\includegraphics[width=0.45\textwidth]{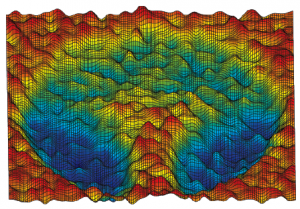}
\caption{Illustration of a complex energy landscape taken from the internet to help with intuition. Sampling (noisy gradient descent) will spend most of its time in low energy : the stationary distribution is $p(x)= \frac{1}{Z}e^{-E(x)}$}
\label{fig:landscape}
\end{figure}

Glossing over the details, the Langevin sampling algorithm is the discretization of a certain version of the Langevin differential equation and can be writen as  :
\begin{align}\label{eqn:langevin}
x^t=x^{t-1}-\frac{\lambda}{2}\underbrace{\nabla_x E_\theta(x^{t-1})}_{\text{Move down potential}}+\underbrace{\omega^k}_{\text{Random bombardment}}
\end{align}
where $\omega^k\sim N(0,\lambda)$.
It can be shown that this converges to $p_\theta$ (i.e. is a valid MCMC algorithm) in the limit of $t\rightarrow \infty$ and $\lambda\rightarrow 0$ \cite{teh}. For a given $\lambda$, Metropolis-Hastings corrections are needed to account for discretization error : the resulting algorithm is called Metropolis-Adjusted Langevin Algorithm (MALA) \cite{mala}. In practice, these corrections are ignored and Langevin sampling is conceptually treated like a noisy gradient descent on the energy landscape, i.e. it is the approximation of a valid MCMC algorithm. See Figure \ref{fig:landscape} for an example of energy landscape.

\subsection{Generating new data with EBMs}

Note that once we have an EBM that correctly models the data distribution, we can generate new "imagined" data points by starting a sampling chain at a random point (i.e. random noise for images) and let it drop to a low-energy point (i.e. a real-looking image).

\section{Learning EBMs with maximum likelihood}
\label{MLE}

Let us see what happens when we try to learn the parameters of an EBM with maximum likelihood estimation, i.e. minimizing the expected negative log-likelihood of the data
\begin{align*}
\lnll&=-\mathbb{E}_{p_{\text{data}}}[\log p_\theta(x)]\\
&=-\mathbb{E}_{p_{\text{data}}}[\log \frac{1}{Z_\theta}e^{-E_\theta(x)}]\\
&=\mathbb{E}_{p_{\text{data}}}[E_\theta(x)]+\log Z_\theta
\end{align*}
Computing the gradient gives
$$\nabla_\theta \lnll=\mathbb{E}_{p_{\text{data}}}[\nabla_\theta E_\theta(x)]+\nabla_\theta \log Z_\theta$$
where the second term can be rewritten as
\begin{align*}
\nabla_\theta \log Z_\theta &= \frac{\nabla_\theta Z_\theta}{Z_\theta}\\
&=\frac{\nabla_\theta \int e^{-E_\theta(x)}dx}{Z_\theta}\\
&=\frac{\int \nabla_\theta e^{-E_\theta(x)}dx}{Z_\theta}\\
&=\frac{\int e^{-E_\theta(x)}\nabla_\theta (-E_\theta(x))dx}{Z_\theta}\\
&=-\int \frac{e^{-E_\theta(x)}}{Z_\theta}\nabla_\theta E_\theta(x)dx\\
&=-\mathbb{E}_{p_\theta}[\nabla_\theta E_\theta(x)]
\end{align*}

Therefore, we can write the learning gradient as
\begin{align}\label{lnll}
\nabla_\theta\lnll&= \underbrace{\mathbb{E}_{p_{\text{data}}}[\nabla_\theta E_\theta(x)]}_{\text{Positive phase}}\underbrace{-\mathbb{E}_{p_\theta}[\nabla_\theta E_\theta(x)]}_{\text{Negative phase}}\\
&=\nabla_\theta\Big(\underbrace{\mathbb{E}_{p_{\text{data}}}[E_\theta(x)]}_{\mathcal{L}_+}\underbrace{-\mathbb{E}_{p_\theta}[E_\theta(x)]}_{\mathcal{L}_-}\Big)
\end{align}
We can separate the loss in two \textit{phases} of learning called positive and negative. The positive phase consists in minimizing the energy of points drawn from $p_{\text{data}}$ and is straightforward to compute assuming that $E_\theta(x)$ is differentiable (like a CNN). This has the effect of making the data points more likely (i.e. push down on their energy). However, this alone would lead to a flat energy landscape (i.e. 0 everywhere). To counteract this, the negative phase pushes up on the energy of points drawn from the model distribution $p_\theta$ : the energy of the points that are unlikely in $p_{\text{data}}$ will only get pushed up while those that are likely are going to get balanced out by the positive phase and will stay relatively low. \textbf{It is the balancing of these two forces that shapes the energy landscape.}

The naïve way to go about computing the negative phase is to collect samples from $p_\theta$ using MC chains and then make a Monte Carlo estimate of the expectation. The problem is that in practice the MCMC chains mix very poorly on complex data. For example, in the space of pixels, plausible images under the data distribution are separated by big distances of highly unlikely images.

In other words, low energy regions in the energy landscape are often separated by immense mountains of high energy which Langevin chains have no chance of traversing in finite time. Hence the samples produced by a chain are not representative of $p_\theta$ and the expectation is strongly biased. The next section proposes an alternative method for training EBMs that sidesteps this problem.

\section{Contrastive Divergence training}
\label{CD}

First, consider the NLL objective written in terms of the KL-divergence (the constant term corresponding to the entropy of the data distribution is omitted because it does not change the objective):
$$\lnll=\dkl[p_{\text{data}}(x)\|p_\theta(x)]$$
The Contrastive Divergence (CD) objective, introduced by \citealp{experts}, can be expressed as follows
\begin{align*}
\lcd&=\underbrace{\dkl[p_{\text{data}}(x)\| p_{\theta}(x)]}_{\lnll}-\dkl[q_{\theta}^t(x)\| p_\theta(x)]
\end{align*}
where $q_\theta^t$ is the distribution resulting from running $t$ steps of the MC chain \textbf{starting from the data}. Notice that the added term is the negative of another NLL loss : it tries to \textbf{minimize} $\mathbb{E}_{q_\theta^t}[\log p_\theta]$, i.e. push $q_\theta^t$ away from $p_\theta$.

\subsection{Equivalence to NLL}

Because $q^0_\theta(x)=p_{\text{data}}(x)$ and $q^\infty_\theta(x)=p_\theta(x)$, the CD loss is often expressed like
$$\lcd=\dkl[q_\theta^0(x)\| q_{\theta}^\infty(x)]-\dkl[q_{\theta}^t(x)\| q_\theta^\infty(x)]$$

Assuming the MCMC sampling algorithm is valid (the stationary distribution of the chains is $p_{\text{data}}$), $q_\theta^t$ is $t$ step closer to the equilibrium than $q_\theta^0=p_{\text{data}}$ and we are guaranteed that $\lcd$ non-negative. Also the two terms are equal if and only if $q_\theta^t=q_\theta^0=p_{\text{data}}$ which in turn imply that $q_\theta^\infty = p_{\text{data}}$, i.e. $\lnll=0$. Therefore, $$\lcd=0\iff \lnll=0$$
so the two objectives have the same solution, but they get there in different ways. The next sections elaborate on this fact. First we motivate this new loss mathematically and then discuss how to think about it versus the NLL loss.

\subsection{Derivation of the gradient}
\label{CD_derivation}

We can write the gradient of the added term like :
\begin{align*}
&\nabla_\theta \dkl[q_\theta^t\|p_\theta(x)]= \frac{\partial \dkl[q_{\theta}^t\| p_\theta]}{\partial q_\theta^t}\frac{\partial q_\theta^t}{\partial \theta}+\frac{\partial \dkl[q_{\theta}^t\| p_\theta]}{\partial
 p_\theta}\frac{\partial p_\theta}{\theta}\\
&=\frac{\partial}{\partial \theta}\dkl[q_{\theta}^t\| p_{\Omega(\theta)}]+\frac{\partial }{\partial \theta}\dkl[q_{\Omega(\theta)}^t\| p_\theta]\\
&= \frac{\partial}{\partial \theta}\dkl[q_{\theta}^t\| p_{\Omega(\theta)}]+\Big(\underbrace{\mathbb{E}_{q_{\Omega(\theta)}^t}[\nabla_\theta E_\theta(x)]}_{\text{Divergence phase}}\underbrace{-\mathbb{E}_{p_\theta}[\nabla_\theta E_\theta(x)]}_{\text{Negative phase}}\Big)
\end{align*}
where $\Omega(\theta)$ means we consider $\theta$ as constant. The last line is obtained essentially the same way as (\ref{lnll}), replacing $p_{\text{data}}$ with $q_{\Omega(\theta)}^t$. This means that the expectation over $p_\theta$ (the problematic negative phase) cancel in the full loss :
\begin{align*}
&\nabla_\theta \lcd= \nabla_\theta \lnll - \nabla_\theta \dkl[q_\theta^t\|p_\theta(x)]\\
&=\underbrace{\mathbb{E}_{p_{\text{data}}}[\nabla_\theta E_\theta(x)]}_{\text{Positive phase}} \underbrace{-\mathbb{E}_{q_{\Omega(\theta)}^t}[\nabla_\theta E_\theta(x)]}_{\text{Divergence phase}} - \underbrace{\nabla_\theta \dkl[q_{\theta}^t\| p_{\Omega(\theta)}]}_{\text{KL term}}\\
&=\nabla_\theta\Big(\underbrace{\mathbb{E}_{p_{\text{data}}}[ E_\theta(x)]}_{\mathcal{L}_+}\underbrace{- \mathbb{E}_{q_{\Omega(\theta)}^t}[ E_\theta(x)]}_{\mathcal{L}_d}\underbrace{- \dkl[q_{\theta}^t\| p_{\Omega(\theta)}]}_{\lkl}\Big)
\end{align*}
Observe that the new gradient doesn't require us to get any samples from the stationary distribution : what we couldn't do anyway. So one way to view CD is like a neat cancellation trick.

\subsection{Divergence phase vs Negative phase}
\label{CD_intuition}

Notice that the divergence phase is very similar to---and in some sense an approximation of---the negative phase of the NLL. However it has quite a different interpretation, mainly because in the divergence phase the chains always start from the data distribution. First of all, notice that the negative phase comes from the \textbf{minimization} of $\lnll=\dkl[p_{\text{data}}(x)\|p_\theta(x)]$ while the divergence phase comes from the \textbf{maximization} of $\dkl[q_{\Omega(\theta)}^t\| p_\theta]$ in $\lcd$.
 
Instead of pushing up on the model distribution, the divergence phase attempts to stop MC chains from \textit{diverging} (hence the name Contrastive Divergence) from the data distribution after running for $t$ steps (by setting up big mountains on the borders of the data manifold). \textbf{This is in some sense a different way of doing the same thing : balancing out the positive phase (remember discussion in section \ref{MLE}}.

This alternative way of approaching the negative phase is the main way in which CD differs from the NLL objective. Note that both objectives have the same solution. Pragmatically, the CD objective is a slight modification of the NLL objective which reframes the problem so that samples from the stationary distribution are no longer needed.

\subsection{KL term and approximate CD}
\label{klterm}

Notice that the CD formulation introduces another term called the KL term. It tries to stop MC chains from diverging (i.e. maximize $\dkl[q_{\theta}^t\| p_\theta]$) but only with respect to $q_\theta^t$ (i.e. the MC chain). Notice that the divergence term does the same thing but with respect to $p_\theta$.

It is ignored in most implementations of CD because is difficult to compute and because in some cases it has been shown to have negligible impact. In practice the loss that is most often optimized is an approximation of the CD loss where this term is dropped :
$$\lcda = \mathcal{L}_+ + \mathcal{L}_d \approx \lcd$$
It has been shown that this loss doesn't follow the gradient of any fixed function and that it is biased \cite{cdguide,bias}. 

\subsection{Small note on $t$}

The value of $t$ varies a lot between different implementations. The general idea is that large values of $t$ work better but are more expensive. Original implementation by \citealp{experts} uses $t=1$. In implementation we are going to use $t=10$.

\subsection{Spurious minima and persistent initializations}
\label{PCD}

Because classical CD starts the MC chains from the data and runs them only for a few steps $t$, only regions near the data manifold are ever explored (and optimized). Regions far away from the data manifold---that have vanishingly low likelihood under the data---can therefore be assigned very high probability under the model. These points are called \textit{spurious minimas}. 

Persistent CD \cite{pcd} is a variation of the classical algorithm that tries to alleviate this problem. At the beginning of training, the MC chains are initialized on the data points like as usual. However, for the subsequent updates, the initial points are taken to be the result of the previous MC chains. The reasoning is that if the learning rate is not too big, the model doesn't change that much between updates so that it is reasonable to say that (at least for a few updates) the multiple chains count as one big chain, i.e. $$p_{\theta_k}^{2t}(x)\approx p_{\theta_k}^t(p_{\theta_{k-1}}^t(x))$$
In words, running many subsequent chains with persistent initialisation and updates of the parameter in between is approximately equal to running one big chain on the last parameter. Of course the more updates you make the less sensible this approximation becomes. For this reason, it can be a good idea to reset the chains to the data after a certain number of updates.

In practice, PCD has been shown to work well even if the chains are very rarely reset \cite{pcd}.

\section{ConvNet EBMs in practice}
\label{practice}

In this section, we review two recent results : \textit{Implicit Generation and Modeling with Energy Based Models} \cite{du2020} and \textit{Improved Contrastive Divergence Training of Energy-Based Model} \cite{du2021}. They are deep learning contributions that make the training and sampling of ConvNet EBMs work in practice. The first paper is one of the first method to successfully train an EBM on complex images and the second paper improves the method by introducing additional tricks. See Figure \ref{fig:image} for a overview of their framework. We start by explaining why the theoretical foundation of their framework is confused and how it can be clarified. Then we describe and justify the different tricks they use in a more principled way.

\subsection{NLL or CD?}

As the name of the second paper indicates, the authors interpret their framework as being an improved version of CD. However, as we will argue, because of an important design decision they made, their learning procedure is actually optimizing the NLL objective. This distinction is subtle but important when reasoning about the algorithm. 

\subsubsection{Initialization of sampling chains}
\label{confusion}

Taking inspiration from persistent CD, \citealp{du2020} propose to have a reservoir of images from which to sample the initialization of the chains for the divergence phase. When a chain finishes, the final image is put back into the reservoir. The reservoir has a limited size (e.g. 10000 in \citealp{du2021}) and adding a new image to it pushes out the oldest element. 

However, at the beginning of training, \textbf{the buffer is initialized with random noise images}. Moreover, every time the buffer is sampled, the image is replaced with random noise with a small probability ($1\%$ in \citealp{du2021}). 

Because of that and what we discussed in section \ref{CD_intuition}, it doesn't make sense to call this CD. Indeed, starting the chains on the data and optimizing them to not diverge is the central point of CD.

The training scheme proposed is much more in line with the NLL objective stance where the samples are actually supposed to come from the $p_\theta$ (whereas they are only expected to come from $q_\theta^t$ in CD). Indeed, having a persistent reservoir can be seen as a way to improve mixing : starting the chains on random noise encourage the samples to cover more of the energy landscape and starting subsequent chains from previous ending points is approximately like running very long chains (which is required to mix).

However, the more the images are complex, the more the energy landscape in pixel space is complex and irregular and the harder it is for chains to mix.

We now discuss each of the tricks used by the authors and justify them as ways of helping the samples to approach the stationary distribution.

\subsection{Energy function}
\label{energy}

The energy function is implemented by a convolutional neural network connected to a fully connected layer that outputs a scalar value. The specific architecture of the CNN depends on the complexity of the images in the dataset. For complex images like faces, state of the art CNN including residual block and self-attention modules are used.

\subsubsection{Compositional Multi-scale Generation}
\label{multiscale}
\begin{figure*}[h]
\centering
\includegraphics[width=\textwidth]{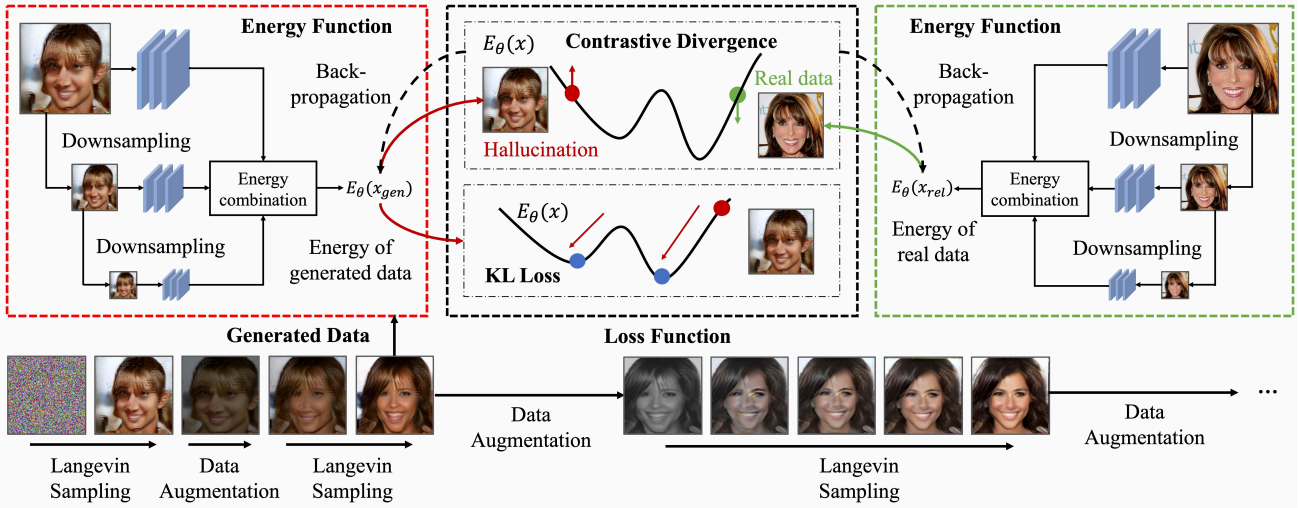}
\caption{Illustration of the overall proposed framework for training EBMs in \citealp{du2021}.}
\label{fig:image}
\end{figure*}
An important architectural component introduced by \citealp{du2021} is the usage of many CNN at different resolutions. The different CNNs act as independent energy functions and each take as input the original image downsampled to certain size (see Figure \ref{fig:image}). Then the final energy of an image is the sum of the CNNs output. Importantly, the sum of many energy function defines the product of their underlying distribution :
\begin{align*}
p_1(x)\cdot p_2(x)&=\frac{e^{-E^{(1)}_\theta(x)}}{Z^{(1)}_\theta}\cdot \frac{e^{-E^{(2)}_\phi(x)}}{Z^{(2)}_\phi}\\
&=\frac{e^{-(E^{(1)}_\theta(x) +E^{(2)}_\phi(x))}}{Z^{(1)}_\theta Z^{(2)}_\phi}
\end{align*}
Therefore, for the energy of an image to be low, only one of the CNN must assign it low energy. In particular, the low resolution CNN can rapidly learn to assign low energy to real images (since they are not trying to model all the details), which means that the sampling will spend more time sensible regions and help the higher resolution CNN to learn.

\subsubsection{Spectral normalization}
\label{spectral}

In \citep{du2020}, authors sometimes found it useful to use spectral normalization on the weights of the CNN. This has the effect of controlling the Lipschitz constant of the energy function, i.e. the maximum absolute value of the slope of $E_\theta(x)$. Doing this makes for a smoother energy landscape : one where sampling is more stable and doesn't fall as often in steep holes. 

\subsection{Sampling}
\label{sampling}

As discussed in section \ref{langevin} the sampling algorithm used is Langevin sampling. However, only using classic Langevin sampling for images leads to especially poor mixing (see end of section \ref{MLE}). Essentially, the sparsity of real images in the space of pixels coupled with the very small steps normally taken by Langevin sampling means MC chains always get trapped in a local modes.

\subsubsection{Step size}
\label{stepsize}

Even if in theory the step size is supposed to be related to with the variance of the noise (see equation \ref{langevin}), in practice both are fixed independently and are considered hyperparameters. Notably the step size is sometimes fixed to high values, between 10 and 1000, while the variance of the noise is usually set to small values like 0.005. High step size are meant to help the sampling chains move faster on the energy landscape, however, this has the effect of making them a poor approximation of the valid MCMC algorithm.

\subsubsection{Data Augmentation Transitions}
\label{augmentation}

In the second paper, the authors propose to periodically perform semantically meaningful transformations to images so that sampling chains can sometimes teleport to very far away regions in pixel space while remaining sensible (see Figure \ref{fig:image}).
Example of transformations are : shift in color, rescalling, Gaussian blur, deformation, etc..

\subsection{Estimation of the KL term}
\label{estimate_kl}

One of the main contributions of  \citealp{du2021} is the addition of auxiliary loss which is meant to estimate the normally ignored KL term. Because of our discussion in section \ref{confusion}, this should not be a good idea. However, the authors made a error in the derivation of this auxiliary loss (flipped a sign) and this resulted in them \textit{minimizing} the KL term, when it technically should be \textit{maximized} (see section \ref{CD_derivation}). As we will see, it turns out that this "doubly wrong" auxiliary loss is a helpful training signal in the context of NLL.

\subsubsection{Optimizing the KL loss}
\label{optimize_kl}

Notice that we can write the gradient of the KL loss as follows :
\begin{align*}
\nabla_\theta \lkl &= \nabla_\theta \Big( -\dkl[q_{\theta}^t\| p_{\Omega(\theta)}] \Big)\\
&=\nabla_\theta \Big(-\mathbb{E}_{q_\theta^t}[\log q_\theta^t]+\mathbb{E}_{q_\theta^t}[\log p_{\Omega(\theta)}]\Big)\\
&=\nabla_\theta\Big(\mathcal{H}(q_\theta^t)+\mathbb{E}_{q_\theta^t}\left[\log \frac{e^{-E_{\Omega(\theta)}(x)}}{Z_{\Omega(\theta)}}\right]\Big)\\
&=\nabla_\theta\Big(\underbrace{\mathcal{H}(q_\theta^t)}_{\lent}\underbrace{-\mathbb{E}_{q_\theta^t}\left[E_{\Omega(\theta)}(x)\right]}_{\lopt}\Big)
\end{align*}
so that the KL loss can be further decomposed in two terms
$$\lkl = \lent + \lopt$$ 
where $\lent$ corresponds to minimizing the entropy of $q_\theta^t$ and $\lopt$ corresponds to pushing up on the points resulting from sampling but \textbf{with respect to the sampling process} $\bm{q_\theta^t}$.

However, because of some sign error made in the derivation, the authors flipped the sign of the objective and therefore optimize the exact opposite. That is, they maximize the entropy of $q_\theta^t$ and minimize the energy of the sampling process. We give more details and explain why and how it makes sense (even if it is wrongly justified).

\subsubsection{Minimizing the energy of sampling}
\label{minimize_sampling}

To minimize the energy of the MCMC samples with respect to the sampling process, one must propagate the gradient through the steps of Langevin sampling and compute second order derivatives. This is a very straightforward task using modern machine learning programming tools since you can just let the gradient flow through the sampling process and let automatic differentiation do the job. However, backpropagating through all the steps is very costly so the authors propose only backpropagating through a few steps. They show that backpropagating through the whole chain leads to very similar results.

\subsubsection{Maximizing the entropy of sampling}
\label{maximize_entropy}

To compute the entropy, the authors propose to use a non-parametric nearest neighbor entropy estimator \cite{ent}. The entropy $\mathcal{H}$ of a distribution $p(x)$ can be estimated through a set $X=x_1,x_2,\ldots,x_n$ of $n$ different points sampled from $p(x)$ as 
$$\mathcal{H}(p(x))= \frac{1}{n}\sum_{i=1}^n\log(n\cdot \text{NN}(x_i,X))+O(1)$$
where $\text{NN}(x_i,X)$ is the minimum distance between $x_i$ and $X$. Therefore, we can \textbf{maximize} the entropy $\mathcal{H}(q_\theta^t)$ by minimizing the loss
$$\lent'=-\mathbb{E}_{q_\theta^t}[\log\text{NN}(x,B)]$$
where $B$ is a set containing some number of past samples from MC chains. By repelling the MCMC samples from each other, we increase entropy, i.e. diversity.

\subsubsection{Why does it work?}
\label{wtf}

In the CD loss, the $\lkl$ loss was meant to make the sampling distribution $q_\theta^t$ diverge from the model distribution $p_\theta$, which makes sense in the context of CD but not so much in the context of NLL. Indeed, in NLL, for the negative phase to be accurate, we need the sampling distribution to converge to the model distribution! 

From this new point of view, minimizing the energy of the sampling process makes sense as an auxiliary objective because it encourages the samples to end up in low points (i.e. which are more likely under $p_\theta$). Also, maximizing the entropy has the effect of diversifying the negative samples, which could all end up in the same hole and give a very biased samples of $p_\theta$. 

Overall, the ``doubly wrong'' auxiliary loss they introduce has the effect of encouraging sampling to closely approximate the stationary distribution $p_\theta$.
\vspace{-2mm}
\section{Experiments}
\label{exp}

We perform some experiments to answer the following questions : 1) does \textit{actual} CD learning lead to good results? 2) what is the effect of better negative sampling on the energy landscape? We perform the experiments on MNIST. 

\subsection{How does \textit{true} CD perform?}

We look at what happens if we modify their framework so that the learning algorithm is \textit{true} CD. To do so, we make a small modification to the initialization buffer : we initialize it with real images and every time a point is sampled from it, it has some chance (e.g. $10\%$) of instead being picked from the dataset (thus keeping the approximation discussed in \ref{PCD} reasonable). No tricks employed in \citealp{du2021} were used. Moreover, we tested if the "right" KL loss could help in that scenario. 

\begin{figure}[H]
\centering
\includegraphics[width=0.4\textwidth]{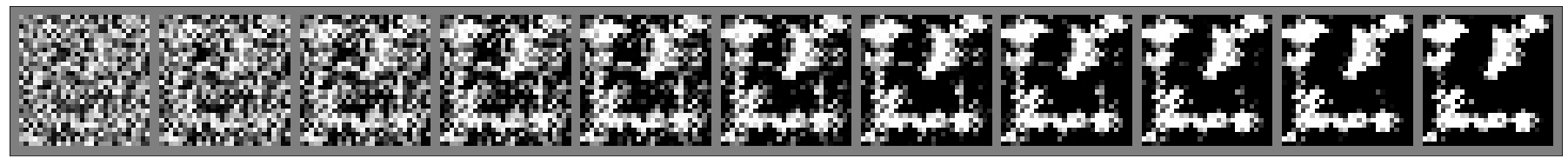}
\includegraphics[width=0.4\textwidth]{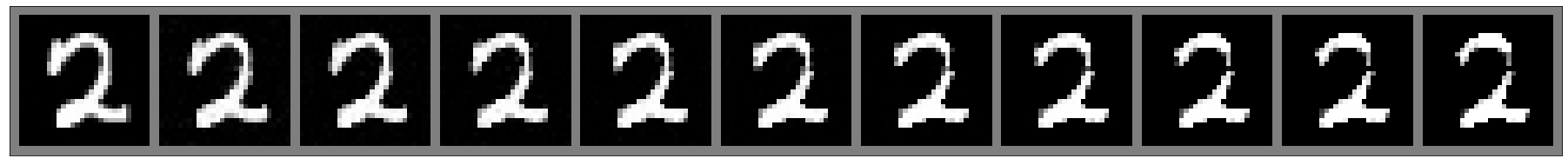}
\includegraphics[width=0.4\textwidth]{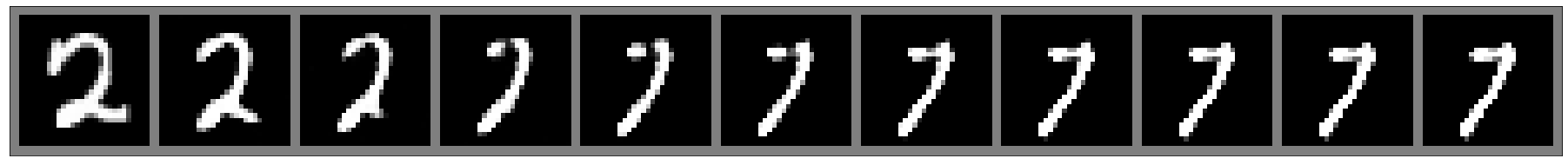}
\caption{Top two are sampling chains on a model trained with true CD. The first starts on random  noise and the second starts on a data point. Bottom image is a chain starting from the same data point but on the EBM trained with the framework in \citealp{du2020}. All are run for the same number of steps (10000) and the same noise (0.005).}
\label{fig:real_init}
\end{figure}

We were not able to make a model trained this way generate satisfying images at test time : while a chain initialized on a real image successfully stays on it (i.e. it doesn't diverge, as the objective encouraged), a chain initialized on random noise always fails to converge to a sensible image (Figure \ref{fig:real_init}). This indicates that the model didn't really learn to fit the data distribution but rather to erect high mountains on the border of the data manifold (as discussed in section \ref{PCD}) and leaving the energy landscape malformed elsewhere. Moreover adding the \textbf{right} KL loss $\lkl$ did not help. However, maybe careful tuning of hyperparameters could lead to better results.
\vspace{-2mm}
\vspace{-2mm}
Overall, we observe that training the EBM with NLL leads to a better energy landscape. For example the last image in Figure \ref{fig:real_init} shows that the chain actually mixes. Also, more pragmatically, if our end goal is to generate new images starting from random points, then it makes sense to make chains start from random points at training (i.e. with NLL).

\subsection{How do the tricks impact the energy landscape?}

\subsubsection{Degeneracy of the energy landscape without the tricks}
Notice that if at test-time we increase the magnitude of the noise, the sampling chains end up in deeper minima. Indeed, noisy gradient descent can more easily escape shallow minima when the noise is high. We observe that when we add enough noise, all chains converge very fast to the same digit (or few digits). This indicate that the energy landscape has a funnel shape and doesn't fit $p_\theta$ well (all digits should be equally likely).
\begin{figure}[H]
\centering
\includegraphics[width=0.4\textwidth]{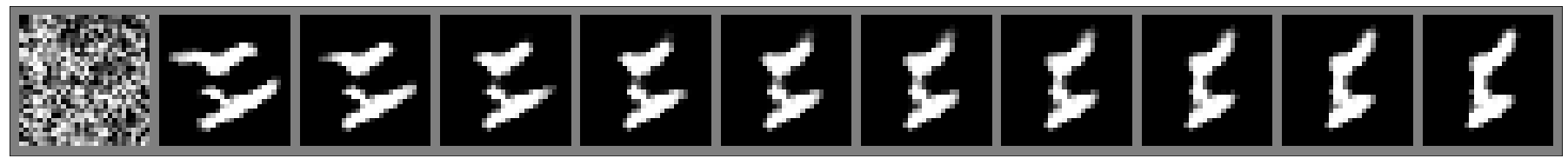}
\includegraphics[width=0.4\textwidth]{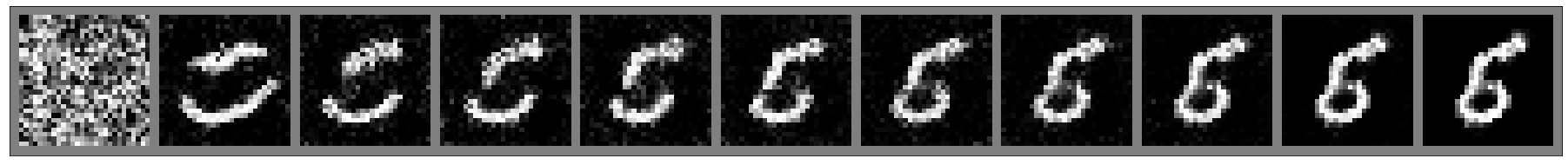}
\includegraphics[width=0.4\textwidth]{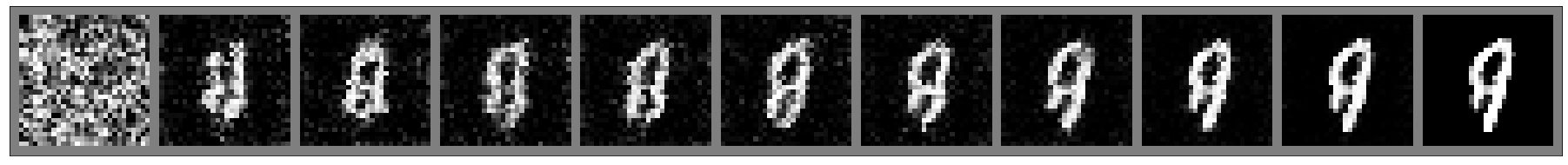}
\includegraphics[width=0.4\textwidth]{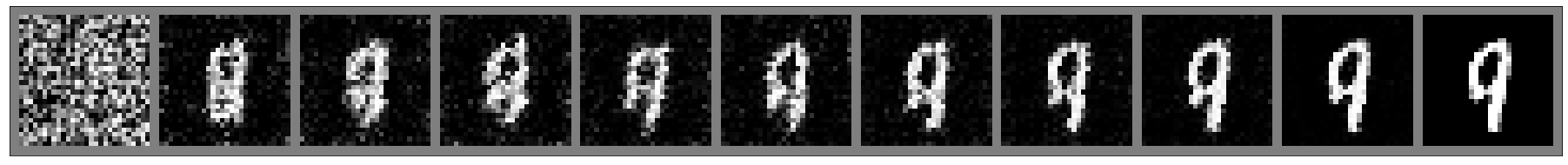}
\caption{Sampling chains with varying noise level. Top three have the same starting point. First two are run for the same number of steps (200) but with (first) low and (second) high noise. Third and fourth are run with high noise for many more steps (10000) but the fourth has a different starting point.}
\label{fig:noise}
\end{figure}
For example, in the model illustrated in Figure \ref{fig:noise}, we observed that with high noise every chain ended up on a 9 no matter the initialization : suggesting 9 is a global energy minimum. We hypothesize that this is because the 9 shares the most features with all other digits : e.g. when we lower the energy of a 1,2,4,7 or 8 the CNN implicitly lowers the energy of all 9.

\subsection{Effect  of better negative sampling}

\begin{figure}[H]
\centering
\includegraphics[width=0.5\textwidth]{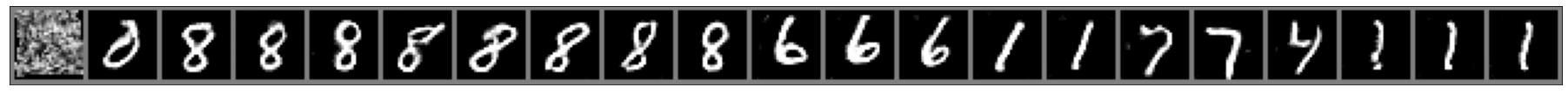}
\includegraphics[width=0.5\textwidth]{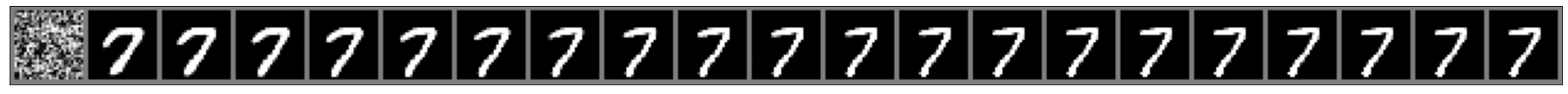}
\includegraphics[width=0.5\textwidth]{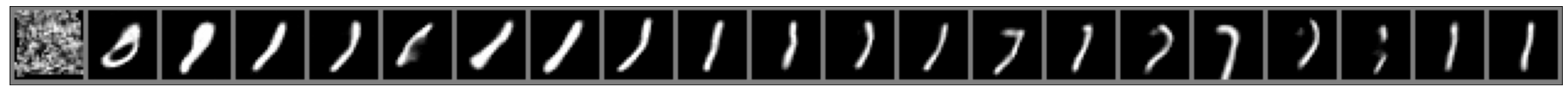}
\caption{Long sampling chains (20000 steps). First is trained and sampled with deformation. Second is trained and sampled without deformation. Third has been trained without deformations but sampled with deformations.}
\label{fig:deform}
\end{figure}

We implemented the procedure of data augmentation transition (section \ref{augmentation}) but with a new type of transformation that is better suited to the MNIST dataset. We used no other tricks (we found that data augmentation was the most impactful). We use elastic deformations, transformations that warp the image in a continuous fashion. In training, every time an image is sampled from the buffer it is slightly warped. Then when generating images at test-time, we apply the transformation every 100 step of Langevin sampling. We observe that this leads to sampling chains that mix much better at test time (i.e. don't fall in a deep hole). See Figure \ref{fig:deform}.

We see from the figure that the deformations don't just help when generating images at test time but more importantly help during training and lead to a better energy landscape. Indeed, at test time, the EBM trained with regular sampling falls in the inescapable deep hole (in this case a "1") even when sampled with deformations.  

This is because during training regular chains don't have time to reach the "1" quickly enough and so the negative phase doesn't balance it properly. More formally, increasing mixing during training means that the negative phase is less biased and the learning gradient of better quality.
\section{Conclusion and future work}
In this work, we offer an in-depth presentation of EBMs sampling and training. Using this knowledge, we showed that \citep{du2020,du2021} have mistaken their training objective for CD while it is in fact NLL. We then discussed different aspect of their training algorithm in light of this insight : in particular, we showed that a mathematical error they made ends up not mattering because of our new interpretation.

We showed --theoretically and experimentally-- that in order to improve current MCMC-based EBM systems, the focus should be on making negative sampling chains mix better, e.g. the tricks in \citep{du2020,du2021}. As we discussed throughout the paper, doing Langevin sampling on a complicated high-dimensional energy landscape (e.g. pixels) necessarily leads to slow mixing because of the small steps. 

Because of that we think that future research should focus on ways to make large steps during the sampling process. This can be done by leveraging knowledge about the structure of the data : for example the \textit{Data Augmentation Transition} trick makes big steps on the energy landscape based on known symmetries of the data (e.g. a crop of a face is still a face). More generally, we think that operating at higher levels of abstraction --where the statistical relationships are lower-dimensional and simpler-- are the key to resolve the inescapable problems of high-dimensional sampling. An exciting recent line of work exploring this idea is the series of papers on GFlowNets \cite{gflow}.

\section*{Acknowledgements}
This work was done as a final project in the class IFT 6269 (Probabilistic Graphical Models) given by Simon Lacoste-Julien in Fall 2021 at Mila.

\bibliographystyle{icml2018}
\bibliography{final.bib}

\section{Implementation details}
\label{details}

\subsection{Dataset}
We used the classic MNIST dataset because it is very simple and doesn't require expensive state-of-the-art CNNs, yet it has enough structure to be interesting to study. It consists in a total of 70000 $28\times 28$ images of hand-written digits, preprocessed to they have a nice standard form.

\subsection{Code}

Our code is based on \cite{tuto}, implementing a basic PyTorch version of the method proposed in the first paper covered \cite{du2020}. We made substantial modification to the code in order to become familiar with it and make it more modular. Additionally, we implemented most of the tricks from \citealp{du2021} in order to test their effectiveness.

\subsection{CNN architecture}

We used the CNN architecture from the code base : a simple 4-layers CNN producing a latent vector of length 128 which is then passed through two linear layers to output a single scalar. We experimented with more powerful architecture but improvement in performance didn't justify the huge different in running time. 

\subsection{Optimizer}

To optimize our model, we used the Adam optimizer \cite{adam} with learning rate $0.0001$ and standard hyperparameters except for the momentum which is set to 0. We also used gradient clipping of $0.1$ to help with training stability.

\end{document}